\documentclass[sigconf]{acmart}
\makeatletter
\renewcommand\@formatdoi[1]{\ignorespaces}
\makeatother
\usepackage{booktabs} 
\usepackage{tabularx}
\setcopyright{none}

\copyrightyear{2018} 
\acmYear{2018} 
\setcopyright{acmlicensed}
\acmConference[ICIAI 2018]{2018 2nd International Conference on Innovation in Artificial Intelligence}{March 9--12, 2018}{Shanghai, China}
\acmPrice{15.00}
\acmDOI{10.1145/3194206.3195587}
\acmISBN{978-1-4503-6345-7/18/03}

\begin{document}
\title{Predictive Liability Models and Visualizations of High Dimensional Retail Employee Data}

\author{Richard R. Yang \footnote{}}
\thanks{This work was completed during the author's undergraduate education at the University of Wyoming.}
\affiliation{%
  \institution{Dept. of Computer Science\\Stanford University}
  \streetaddress{450 Serra Mall}
  \city{Stanford} 
  \state{California} 
  \postcode{94305}
}
\email{richard.yang@cs.stanford.edu}

\author{Mike Borowczak}
\affiliation{%
  \institution{Dept. of Computer Science\\ University of Wyoming}
  \streetaddress{1000 E. University Ave}
  \city{Laramie} 
  \state{Wyoming} 
  \postcode{82070}
}
\email{mike.borowczak@uwyo.edu}


\begin{abstract}
Employee theft and dishonesty is a major contributor to loss in the retail industry. Retailers have reported the need for more automated analytic tools to assess the liability of their employees. In this work, we train and optimize several machine learning models for regression prediction and analysis on this data, which will help retailers identify and manage risky employees. Since the data we use is very high dimensional, we use feature selection techniques to identify the most contributing factors to an employee's assessed risk. We also use dimension reduction and data embedding techniques to present this dataset in a easy to interpret format.

\end{abstract}


\maketitle

\section{Introduction}
In 2016, the retail industry suffered an estimated loss of \$44 billion USD due to theft, employee dishonesty, and administrative accounting errors \cite{NRF}. Experts attribute 65\% to 80\% of retailers' total loss to employee dishonesty \cite{schulte2012how}. Since revenue loss from employee dishonesty significantly affects retailers, we partnered with a major U.S.-based retailer and investigate their current method of assessing employee risk. Our retailer's risk evaluation system requires a complicated, manually hand-tuned weighting process done by human analysts \cite{sysrepublic}. The result of this risk assessment system is a single numeric risk factor value that determines an employee's risk to the store.

In discussions with the authors, an executive of our partner retailer expressed concern that the retail industry, while a data rich environment, has many data sources that are disconnected and under-analyzed. This is an ideal opportunity to utilize the vast amount of data that can be provided by retailers to create a better system of determining employee risk. Our partner retailer provided us with a dataset employee statistics from stores across the U.S., and the manually assessed risk factor pertaining to those employees. In our work, we investigate 1) an automated approach in analyzing employee risk using machine learning models, 2) the features that primarily identify high risk employees, and 3) a method to represent employee data such that it is easily interpretable by employee supervisors. 

First, we train a variety of regression models to learn the relationship between certain employee statistics and their assessed risk factor. The trained regression models allow us to predict the risk of other employees using the same statistics.

Second, we use recursive feature elimination on the trained models to identify the top contributing features to an employee's assessed risk. By identifying the most important features, we can reduce the amount of data that needs to be collected in order to assess employees.

Lastly, real-world data from retail (and other) industries are rarely contained in two or three dimensions, so it can not be visualized through typical means of plotting. Even if the data could be easily visually represented, it is unlikely that each dimension would be scaled based on contextual meaning of the underlying data source. These real-world, high-dimensional datasets contain complex relationships between features and dimensions but lack straightforward methods to transform and visualize these relationships. This aspect makes understanding the data an unintuitive and complicated challenge for analysts. However, it is possible to extract meaningful low-dimension features from a high dimensional dataset by using dimension reduction and data embedding methods.

\section{Methodology}
\subsection{Data Description}
Our retail partner provided us with a real 31-dimensional dataset of cashier activity from their stores across the U.S. Each entry of the dataset consists of the cashier's identification number, store identification number, and twenty-eight other features. The data was recorded by the corporate retail chain from their daily operations. The database is organized into sheets by store region (Denver, Eastern, Houston, Inter-mountain, Northern California, Portland, Seattle, Southern California, Southern, Southwest), with around 100 entries for each region. Combined, there were a total of 991 entries for the entire set. Note that this is only a subset of the full data available from the retailer. Also provided with each entry was a risk factor manually determined and calculated by a risk assessment consulting firm using existing methodology.


\subsection{Data Processing}
\begin{figure}[!t]
\centering
\includegraphics[width=3.5in]{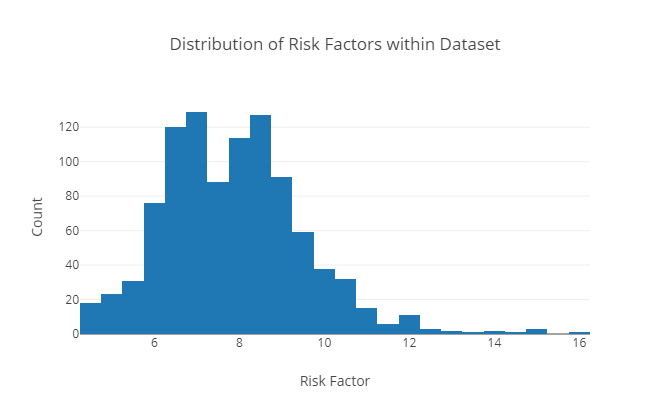}
\caption{Distribution of risk factor for our dataset}
\label{fig_histogram}
\end{figure}

The provided dataset is in the format of a rich-text Microsoft Excel workbook. To clean our data, we strip the data values from its formatting and export it to a CSV file. Then, we read the file into a (991,31) matrix. We remove two statistically unimportant features from the dataset: the store ID and cashier ID. We then split the risk factor column from the rest of the dataset to use as labels for our regression models. The distribution of of risk factors is shown in figure \ref{fig_histogram}. The features are not evenly scaled, so we perform data standardization such that the dataset has zero mean and unit variance. Our data is split into 90\% training set and 10\% testing set.

\subsection{Regression Models}
We train a variety of regression models on all 28 features to predict a numeric risk factor. In this section, we give a high level overview of the learning algorithms we use. 

The first model we train is K Nearest Neighbors (KNN). The training process for KNN is simple: the model stores the data and the respective labels. When predicting the new risk factor, the model samples the $k$ closest training data points, and returns the mean of the points as the prediction.

The second class of models we optimize is linear regression, where the models approximate the regression function as a weighted linear combination of the input features. We fit the data to ordinary least squares linear regression, lasso regression, and ridge regression. In ordinary least squares, we optimize our model to minimize the residual sum of squares. In lasso regression, we add a L1 regularization to the minimization objective. Similarly, in ridge regression, we add a L2 regularization to the minimization objective. The regularization factors are used to penalize the model on selecting insignificant features, and thus reducing the amount of over-fitting on the training set.

The third class of models we optimize are support vector machines for regression (SVRs) with two types of kernels: linear function and radial basis function (RBF). In the case of SVMs for classification, the models are trained to identify a hyperplane separating the data points. For SVRs, we instead train the model to estimate a function such that the data points are within an $\epsilon$ from the function. The linear kernel maps input to an inner product of the samples, and the RBF kernel maps input to an exponential function of the squared Euclidean distance between the samples. We chose the linear kernel for faster optimization, and the RBF kernel for better predictive power.

The fourth class of models we optimize is the decision tree and random forest algorithm. In decision tree learning, we fit the data to a model represented by a tree structure with conditions as nodes. The model predicts the risk factor by traversing the tree and outputting the value of the leaf node. In the random forest algorithm, multiple different decision trees are generated and fit to the training data. The algorithm predicts the risk factor by propagating the input features to all decision trees in the ensemble, and taking the mean value across all predictions.

We use the Scikit-Learn Python library for the implementation of the algorithms mentioned here \cite{scikit-learn}. 

\subsection{Feature Selection}
We train our models with 28 features, but the complex dynamics the models learn may not be intuitive for humans to understand. In the case where a manager is reviewing the employee records, it is not feasible to manually sift through 28 columns of data. To help make this process easier, we use feature selection techniques to identify the top contributing features to an employee's assessed risks. 

Recall that we optimize linear regression models, which predict using a weighted linear combination of the input features. The parameters that the models learn are coefficients in the regression function, and we leverage these to select and interpret the features. Since our data has been standardized, then by intuition, the most important features should have the highest coefficients in our learned model. In the results section, we report the top two weighted features from each of our linear models.

A more powerful method to identify important features is the recursive feature elimination (RFE) algorithm. RFE is akin to running a greedy search to find the subset of features that perform the best with the model. We train the models using only a subset of the features, and recursively prune the least contributing feature until we have only two features remaining. We run RFE in conjunction with our linear regression models, SVR, and random forest.

\subsection{Visualization}





With our feature selection techniques, we narrow down the feature space from 28 to 2. With a 2-dimensional feature space, we can plot and visualize the dataset. However, the relationship between the 26 other features would be lost, and the resulting plot is not a good representation of the data.

Our first visualization technique is to perform principal component analysis (PCA) on our data. PCA identifies a directional axis that results in the highest variance of the data, called the first principal component \cite{shealy2016dimensionality}. Then, it linearly projects the high dimensional dataset into a lower dimensional space described by the first principal component and direction orthagonal to it. The result is the data transformed into two dimensions that can be visualized through plotting.

Our second method of visualization is t-Distributed Stochastic Neighbor Embedding (t-SNE). t-SNE creates a two-dimensional embedding of the original data, where the points' location is selected by a probability distribution proportional to a similarity measure of the two data points in the high dimensional space. The optimization step for t-SNE is to minimize the KL-divergence between the probability distribution of the low-dimensional embedding and the original data \cite{maaten2008visualizing}.

\section{Results and Analysis}
\subsection{Regression Models}
For performance evaluation, we report the mean squared error, mean absolute error, and coefficient of determination ($R^2$ score) on the test set for each model. Mathematically, the MAE is the average of $|\text{Prediction} - \text{True Value}|$ across all examples, the MSE is the average of $(\text{Prediction} - \text{True Value})^2$ across all examples,  and $R^2$ is $(1 - \text{Mean Residual Sum of Squares}) / \text{MSE}$. The evaluation across all models is shown in table \ref{tab_reg}.

For KNN, a common practice is to select the number of neighbors $k$ as the square root of the number of features. We tune our KNN model by varying $k$ as a hyperparameter and select the best performing number.

In addition to training our regularized linear regression models (lasso and ridge), we also perform tuning by adjusting the hyperparameter $\alpha$. This hyperparameter is the regularization penalty factor, so the model is penalized more for larger norms for larger $\alpha$ values.

\begin{table}[!htb]
\caption{Comparison of mean average error, mean squared error, and coefficient of determination across all regression models} \label{tab_reg}
\setlength\tabcolsep{0pt} 
\smallskip 
\begin{tabular*}{\columnwidth}{@{\extracolsep{\fill}}rcccr}
\toprule
  Models  & MAE& MSE & R2 \\
\midrule
KNN                 & 0.8455          & 1.173           & 0.458           \\ 
Linear Regression   & 0.8407          & 1.68            & 0.2239          \\ 
Lasso Regression    & 0.8567          & 1.26            & 0.4178          \\ 
Ridge Regression    & 0.8218          & 1.462           & 0.3243          \\ 
SVR (Linear Kernel) & 0.8239          & 1.78            & \textbf{0.1776} \\
SVR (RBF Kernel)    & \textbf{0.6816} & \textbf{0.7948} & 0.6328          \\ 
Decision Tree       & 0.9744          & 1.68            & 0.2239          \\ 
Random Forest       & 0.7398          & 0.9143          & 0.5776  		   \\
\bottomrule
\end{tabular*}
\end{table}

The risk factors in our dataset range from 4 to 16. Our top performing model, support vector regression using an RBF kernel, achieves a desirable MAE of 0.6816, and MSE of 0.7948. Under intuition, we believe that the linear models did not perform as well because the high dimensional data is not linearly correlated. With the decision tree algorithm, there exists too many branching factors for one single tree to be a good representation of the dataset. However, in random forest, we use an ensemble of decision trees and saw improved performance.

\subsection{Feature Selection}
First, we report the top two features as ranked by our linear regression models' coefficients. For linear regression: "Department Sales Item Count" and "Department Refunds Item Count". For lasso regression: "Number of Sales Per Day and Number" of "Item Voids". For ridge regression: "Department Sales Total Amount" and "Department Sales Item Count". There are many possible explanations for why these features were selected (e.g. the model may have learned that a large number of item voids is generally representative of high risk), and we leave it open to interpretation by the retail managers experienced in this field.

\begin{figure}[t]
\centering
\includegraphics[width=3.5in]{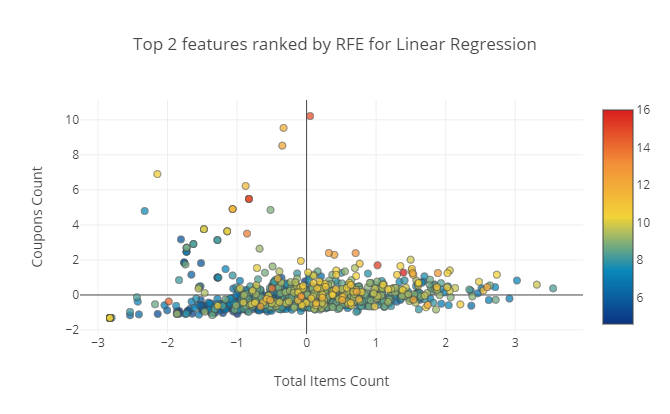}
\includegraphics[width=3.5in]{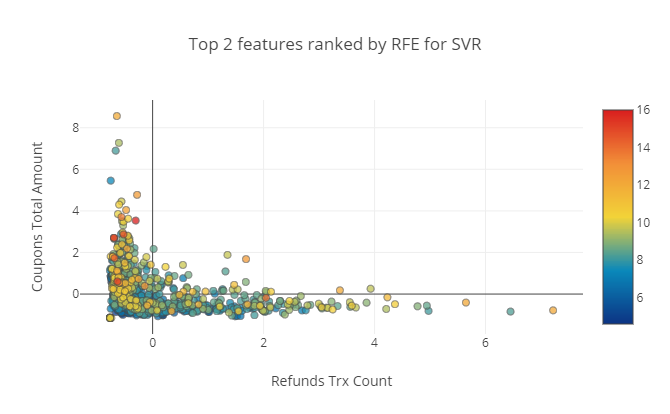}
\includegraphics[width=3.5in]{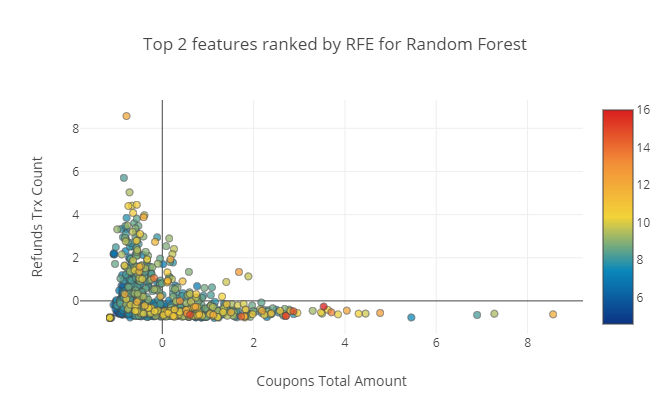}
\caption{Plots of the data using only the top 2 features ranked by RFE for Linear Regression, Support Vector Regression, and Random Forest. The points are colored based on the risk factor.}
\label{fig_rfe}
\end{figure}

Next, we report and plot the top two features selected by the recursive feature elimination algorithm on certain models shown in figure \ref{fig_rfe}. We use RFE to identify the most important features contributing to an employee's assessed risk. This information is useful to human supervisors that are looking for areas to help their employees improve in. However, in terms of visualization, these plots do not reveal much information about the dataset. Most of the points are centralized into one cluster with some minor outliers.

\subsection{Visualization}
The main challenge with visualizations on this dataset is assigning an objective evaluation of how useful the plots are. Qualitatively, a useful plot would allow the reader to distinguish areas (clusters) of varying risk. Quantitatively, the closest evaluation metric is the silhouette score, which is typically used to assess clustering algorithms. The silhouette score is a measurement of similarity between a data point and points in the same cluster and points outside of the cluster \cite{mwangi2014visualization}. 

We first have to choose the number of clusters in the data in order to evaluate the silhouette score, since it is a measure of clustering performance. While it is possible to choose an arbitrary number of clusters, we methodically run k-Means on the original dataset with varying $k$ values from 2 to 10 and evaluate it with the silhouette score. We obtain the highest silhouette score with $k=2$, so we evaluate our visualization methods using the silhouette score and two classes.

\subsubsection{PCA}

\begin{figure}[t]
\centering
\includegraphics[width=3.5in]{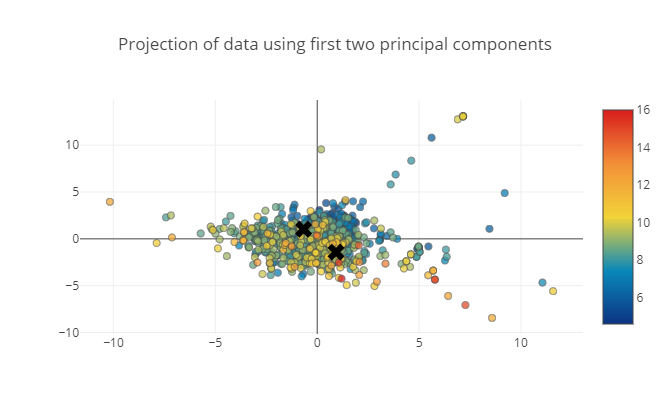}
\caption{Plots of the data using two principal components as identified by PCA. The X's mark the location of the centroids calculated by k-Means.}
\label{fig_pca}
\end{figure}

After performing PCA on the dataset to identify two principal components, we show the resulting projections in figure \ref{fig_pca}. Qualitatively, this visualization does not reveal much meaningful relationships between the data points. Although PCA is a fast algorithm to run on this dataset, it is not the most meaningful tool to use for our purposes.

\subsubsection{t-SNE}
\begin{figure}[t]
\centering
\includegraphics[width=3.5in]{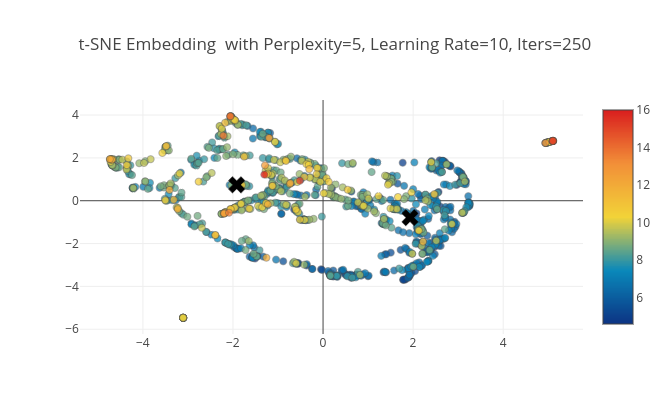}
\includegraphics[width=3.5in]{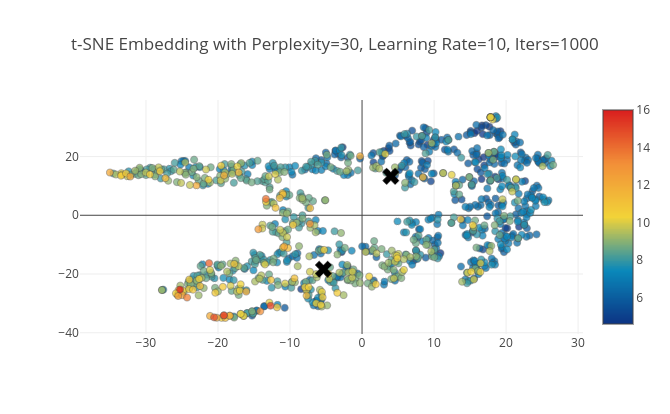}
\includegraphics[width=3.5in]{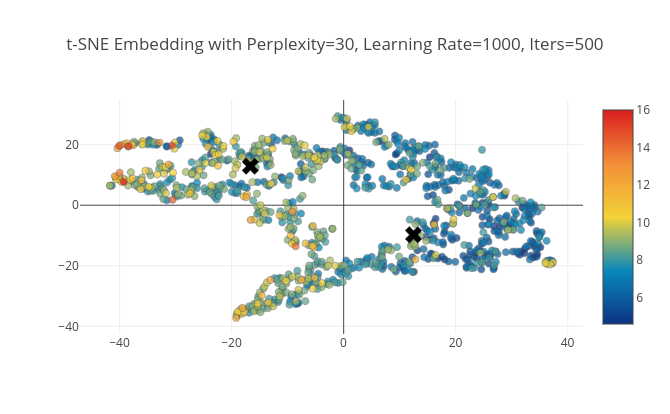}
\caption{Top three t-SNE embeddings evaluated by silhouette score. The X's mark the location of the centroids calculated by k-Means. These plots are both qualitatively and quantitatively better than previous approaches.}
\label{fig_tsne}
\end{figure}

\begin{figure}[t]
\centering
\includegraphics[width=3.5in]{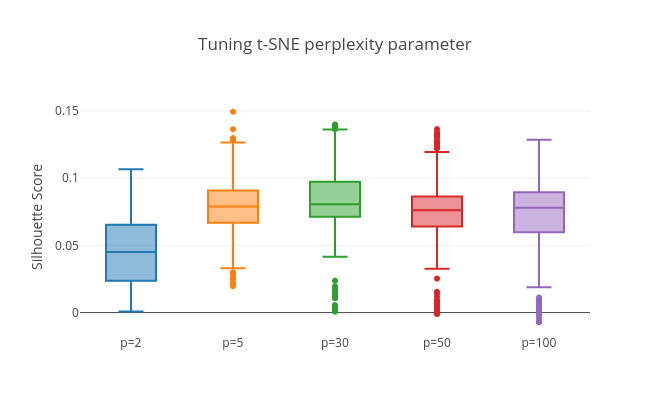}
\includegraphics[width=3.5in]{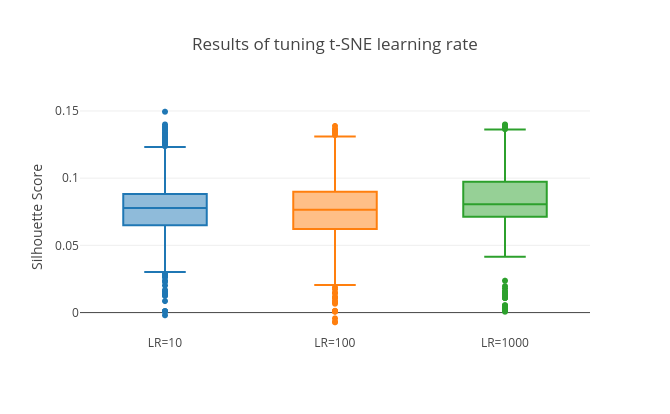}
\caption{Box plots depicting the effect of hyperparameter tuning on t-SNE.}
\label{fig_tsne_hyperparam}
\end{figure}

The optimization objective of t-SNE is non-convex, and t-SNE will produce very different visualizations based on three hyperparameter settings that must be tuned \cite{scikit-learn} \cite{wattenberg2016how} . The first hyperparameter is the perplexity, which is a balance between local and global aspects of the data. The second is the number of iterations to run the algorithm for. The last is the optimization learning rate. To tune the ideal hyperparameter settings, we perform grid search using a log-scale with the silhouette score as the metric. To highlight the importance of the hyperparameters, we report the mean silhouette score and confidence intervals for each setting in figure \ref{fig_tsne_hyperparam}. The top-performing t-SNE embeddings are shown in figure \ref{fig_tsne}. The training and tuning process for t-SNE is significantly longer than the time required to run PCA on the dataset but yields much more useful visualizations.

\subsubsection{Summary}
An overview of the three different visualization techniques we use is shown in table \ref{tab_ss}. We show that applying PCA to this dataset is not sufficient for the visualization task we proposed, as it is our worst performing method judged by the silhouette score. Although visualizing the top two features selected using RFE yields high silhouette scores, it is also not sufficient if we want to capture the full essence of the dataset. We report that a tuned t-SNE embedding is the best performing method, both quantitatively and qualitatively. 

\begin{table}[]
\centering
\caption{A comparison of visualization techniques using the silhouette score as measure}
\label{tab_ss}
\begin{tabular}{lll|l}
\hline
\multicolumn{3}{c|}{Visualization Method}                                                            & \multicolumn{1}{c}{Silhouette Score} \\ \hline
\multicolumn{3}{l|}{Top 2 RFE Features for Lin Reg}                                                  & 0.06615                              \\
\multicolumn{3}{l|}{Top 2 RFE Features for SVR}                                                      & 0.12499                              \\
\multicolumn{3}{l|}{Top 2 RFE Features for Random Forest}                                            & 0.12209                              \\
\multicolumn{3}{l|}{Principal Component Analysis}                                                    & 0.06615                              \\ \cline{1-3}
\multicolumn{3}{l|}{Top 3 t-SNE Settings}                                                            & \multicolumn{1}{c}{}                 \\
\multicolumn{1}{c}{Perplexity} & \multicolumn{1}{c}{Learning Rate} & \multicolumn{1}{c|}{Iterations} &                                      \\ \cline{1-3}
5                              & 10                                & 250                             & \textbf{0.14943}                     \\
30                             & 10                                & 1000                            & 0.13992                              \\
30                             & 1000                              & 500                             & 0.13929                              \\ \hline
\end{tabular}
\end{table}

\section{Conclusion}
In this work, we train and optimize multiple regression machine learning models to predict the risk of a retail employee given some work performance. These automated models are an improvement to existing systems that require manual, hand-tuned assessment of employee risk. We achieve a satisfiable mean absolute error using support vector regression with an radial basis function kernel. From the regression models we train, we also extract coefficients representative of the most important features in the dataset using the recursive feature elimination algorithm. We hope that retail store managers can utilize this information to focus on areas of improvement for their employees. Lastly, we present this data in a easy to interpret medium for humans. Initially a 28-dimensional dataset, we represent this data in a interactive 2-dimensional plot using a dimension reduction and data embedding techniques. We conclude that t-SNE is superior in terms of data embedding over the traditional PCA method. Our hope is that retail store managers can easily interact with our visualizations to rank their employees, rather than sifting through a large amount of number data. 
\section*{Acknowledgement}
The authors would like to thank Kathleen Smith - Vice President of Loss Prevention at Safeway Inc., and the University of Wyoming Cybersecurity Education and Research (CEDAR) Center.

\bibliographystyle{ACM-Reference-Format}
\bibliography{references} 

\end{document}